\documentclass{amia}
\usepackage{graphicx}
\usepackage[labelfont=bf]{caption}
\usepackage{subcaption}
\usepackage[superscript,nomove]{cite}
\usepackage{url}
\usepackage{xspace}

\usepackage{color}
\usepackage{colortbl}
\usepackage{array}
\usepackage{bibspacing}
\usepackage{booktabs}
\usepackage{float}

\makeatletter
\DeclareRobustCommand\onedot{\futurelet\@let@token\@onedot}
\def\@onedot{\ifx\@let@token.\else.\null\fi\xspace}

\def\eg{\emph{e.g}\onedot} 
\def\ie{\emph{i.e}\onedot}

\def\etal{\emph{et al}\onedot}
\makeatother

\newcommand{\myemph}[1]{\textbf{\textit{#1}}}
\newcommand{\cnnmodel}{\emph{CNN model\xspace}}
\newcommand{\grassmannmodel}{\emph{Grassmann model\xspace}}
\newcommand{\multitaskmodel}{\emph{Multi-task model\xspace}}

\begin{document}

\title{A multi-task deep learning model for the classification of Age-related Macular Degeneration}

\author{Qingyu Chen, PhD$^{1}$\footnote{These authors contributed equally to this work.}, Yifan Peng, PhD$^{1*}$, Tiarnan Keenan, BM BCh, PhD$^{2}$, Shazia Dharssi$^{2}$, Elvira Agr\'on, MS$^{2}$, Wai T. Wong, MD$^{2}$, Emily Y. Chew, MD$^{2}$, Zhiyong Lu, PhD$^{1}$}

\institutes{
    $^1$National Center for Biotechnology Information (NCBI), National Library of Medicine (NLM), National Institutes of Health (NIH), Bethesda, Maryland, United States;\\
    $^2$National Eye Institute (NEI), National Institutes of Health (NIH), Bethesda, Maryland, United States;
}

\maketitle

\noindent{\bf Abstract}

\textit{Age-related Macular Degeneration (AMD) is a leading cause of blindness. Although the Age-Related Eye Disease Study group previously developed a 9-step AMD severity scale for manual classification of AMD severity from color fundus images, manual grading of images is time-consuming and expensive. Built on our previous work DeepSeeNet, we developed a novel deep learning model for automated classification of images into the 9-step scale. Instead of predicting the 9-step score directly, our approach simulates the reading center grading process. It first detects four AMD characteristics (drusen area, geographic atrophy, increased pigment, and depigmentation), then combines these to derive the overall 9-step score. Importantly, we applied multi-task learning techniques, which allowed us to train classification of the four characteristics in parallel, share representation, and prevent overfitting. Evaluation on two image datasets showed that the accuracy of the model exceeded the current state-of-the-art model by $>10\%$. 
}

\section*{Introduction}

Age-related Macular Degeneration (AMD) is responsible for 9\% of blindness worldwide but is the leading cause in developed countries. The number of patients diagnosed with AMD worldwide is projected to increase to 288 million by 2040\cite{wong2014global}. Based on clinical features, the disease is classified into early, intermediate, and advanced stages\cite{ferris2013clinical}. Advanced AMD, which is often associated with severe visual loss, can occur in two forms: geographic atrophy (or `dry') and neovascular (or `wet') AMD.

To determine the severity of non-advanced AMD, color fundus photographs are graded on a scale of 1--9. Additional steps on the scale (10--12) are sometimes used to grade advanced AMD\cite{group2005age}. However, human grading using this severity scale requires trained expert graders and is highly time-consuming\cite{burlina2017automated}, and usually only performed at dedicated reading centers. This has limited the use of this severity scale to a research tool, rather than as part of the clinical care of patients\cite{group2005simplified}; even so, researchers need access to reading centers, which can be very expensive in the case of large studies. Therefore there is currently an unmet need for algorithms that can perform automated grading of AMD severity from color fundus photographs, which will be helpful both for research  decision-making in both research and clinical practice.

Early retinal image classification systems of color fundus photographs had adopted traditional machine learning with human-engineered features\cite{dalal2005histograms}. Subsequently, later systems used deep learning methods as feature extractors\cite{burlina2017comparing}. Deep learning has revolutionized the computer vision domain\cite{krizhevsky2012imagenet}, and has become the state-of-the-art approach for medical image classification\cite{litjens2017survey,kim2016deep,margeta2017fine,wang2017chestx,wang2018tienet}. To date, several groups have applied deep learning methods to AMD severity classification using color fundus photographs\cite{burlina2017comparing}. The state-of-the-art method, recently reported by Grassmann \etal\cite{grassmann2018deep}, treated this task as an image classification problem. The approach of these authors was to use six individual models (AlexNet\cite{krizhevsky2012imagenet}, GoogLeNet\cite{szegedy2015going}, VGG\cite{simonyan2014very}, Inception-V3\cite{szegedy2016rethinking}, ResNet\cite{he2016deep} and Inception-ResNet-V2\cite{szegedy2017inception}), each trained from scratch. Each model directly predicted the step on the Age-Related Eye Disease Study (AREDS) severity scale 1-12 from the color fundus photograph (\ie, six separate predictions), then a random forest approach was employed to combine the predictions into one overall prediction.

However, direct classification of images into the AREDS severity scale does not reflect normal human grading practice. In reading centers, rather than grade color fundus photographs directly, certified graders first calculate individual scores for four separate AMD characteristics (drusen area, geographic atrophy, increased pigment, and depigmentation), then combine the scores for these four characteristics into the 9-step non-advanced AREDS severity scale\cite{group2005age}. The method for combining these scores into the AREDS severity scale is shown in Table~\ref{tab:amd}. For example, for one color fundus photograph, if the drusen area is 1 and geographic atrophy, increased pigment, and depigmentation are all absent (assigned 0), then the AREDS severity scale is defined as 2 for that image. Graders also check separately for additional characteristics of advanced AMD (scale 10-12), and revise the severity scores if necessary. Hence, a deep learning approach that predicts the overall AREDS severity score directly (as in Grassmann~\etal\cite{grassmann2018deep}), without these intermediate steps, may have lower transparency and decreased information content for research and clinical purposes\cite{madumal2018towards,doshi-velez2017towards}.
\begin{table}[H]
\vspace{1em}
\small
\newcolumntype{x}[1]{>{\centering\arraybackslash\hspace{0pt}}p{#1}}
\definecolor{color1}{RGB}{255,255,255}
\definecolor{color2}{RGB}{227,228,229}
\definecolor{color3}{RGB}{167,196,229}
\definecolor{color4}{RGB}{188,220,217}
\definecolor{color5}{RGB}{254,251,232}
\definecolor{color6}{RGB}{254,248,196}
\definecolor{color7}{RGB}{248,230,192}
\definecolor{color8}{RGB}{251,236,243}
\definecolor{color9}{RGB}{241,200,192}

\newcommand{\cellone}{\cellcolor{color1}{1}}
\newcommand{\celltwo}{\cellcolor{color2}{2}}
\newcommand{\cellthree}{\cellcolor{color3}{3}}
\newcommand{\cellfour}{\cellcolor{color4}{4}}
\newcommand{\cellfive}{\cellcolor{color5}{5}}
\newcommand{\cellsix}{\cellcolor{color6}{6}}
\newcommand{\cellseven}{\cellcolor{color7}{7}}
\newcommand{\celleight}{\cellcolor{color8}{8}}
\newcommand{\cellnine}{\cellcolor{color9}{9}}

\caption{AREDS Severity Scale scores 1 to 9, defined by graders from four categories: geographic atrophy (0/1, \ie, absent/present), increased pigment (0/1, \ie, absent/present), depigmentation (graded 0-3), and drusen area (graded 0-5). The final AREDS Severity Scale score (steps 1-9, shown shaded in different colors) is defined by the combination of findings from these four categories.\label{tab:amd}}
\centering
\begin{tabular}{lx{3em}x{3em}x{3em}x{3em}x{3em}x{3em}}
\hline
& \multicolumn{6}{c}{Pigment abnormalities}\\
\cline{2-7}
& \multicolumn{6}{|c|}{}\\
Geographic atrophy & 0 & 0 & 0 & 0 & 0 & 1\\
Increased pigment & 0 & 1 & -- & -- & -- & --\\
Depigmentation & 0 & 0 & 1 & 2 & 3 & --\\
Drusen area\\
\hline
0 & \cellone{} & \celltwo{} & \celltwo{} & \cellfour{} & \celleight{} & \cellnine{}\\
1 & \celltwo{} & \cellfour{} & \cellfour{} & \cellfour{} & \celleight{} & \cellnine{}\\
2 & \cellthree & \cellfour{} & \cellfour{} & \cellfive{} & \celleight{} & \cellnine{}\\
3 & \cellfour{} & \cellfive{} & \cellfive{} & \cellsix{} & \celleight{} & \cellnine{}\\
4 & \cellfive{} & \cellsix{} & \cellsix{} &  \cellseven{} & \celleight{} & \cellnine{}\\
5 & \cellsix{} & \cellseven{} &  \cellseven{} & \celleight{} & \celleight{} & \cellnine{}\\
\hline
\end{tabular}
\vspace*{-1em}
\end{table}

To address these potential criticisms, we designed a novel deep learning approach, which mirrors more closely the way that human graders in reading centers perform grading according to the AREDS severity scale. We have focused on the classification of non-advanced AMD (scale 1-9), because this was the original intention of the AMD severity scale (\ie, to predict risk of progression to advanced AMD).

Intuitively, we can design four deep learning models (each of which is responsible for the classification of an individual characteristic) and train them separately, which is called single-task learning. Indeed, we previously designed DeepSeeNet, a deep learning model for the classification of AMD (at the patient level) that uses single-task learning\cite{peng2018deepseenet}. However, these four variables are related; training separately may cause the model not to benefit from shared features (from the other variables) and may overfit to specific variables. Instead, we created a multi-task deep learning model that trains the classification of the four characteristics simultaneously. Multi-task learning allowed us to exploit the similarities and differences between the four characteristics, via the shared deep learning model layers, and also allowed us to reduce losses from specific tasks\cite{zhang2017survey}. While multi-task learning has been successfully used in computer vision\cite{zhang2014facial,ranjan2017hyperface} and natural language processing applications\cite{subramanian2018learning,soegaard2016deep}, to the best of our knowledge, this is the first report that has employed a multi-task learning model in AMD classification.

For the evaluation of our model’s robustness and generalizability, two datasets were used. One was from the AREDS, which includes a large dataset of color fundus photographs (publicly available, on request). The other was a newly created dataset, from the AREDS2. The results of our experiments, using these two datasets via 5-fold cross-validations, demonstrated that our model performed consistently better than the state-of-art model (from Grassmann \etal\cite{grassmann2018deep}). In particular, the F1-score and accuracy were $\sim 5\%$ and $\sim 4\%$ higher in our model, respectively, in terms of absolute differences. Furthermore, the model explains the final classification outcome by providing the intermediate results regarding the four separate characteristics, which may be useful for research or clinical purposes.

Our deep learning model and data partition are publicly available\footnote{it will be released on AMIA informatics 2019.}. To the best of our knowledge, this is the first study in the field of AMD severity scale classification to make these elements publicly available. The goal is to allow for transparency and reproducibility of this approach, so that this model may serve as a benchmark method to allow for further advancement of state-of-the-art techniques.

\section*{Methods}

\subsection*{Architecture of AMD characteristics}

\begin{figure}[H]
\centering
\includegraphics[width=.7\textwidth,trim={4cm 2cm 4cm 2cm},clip]{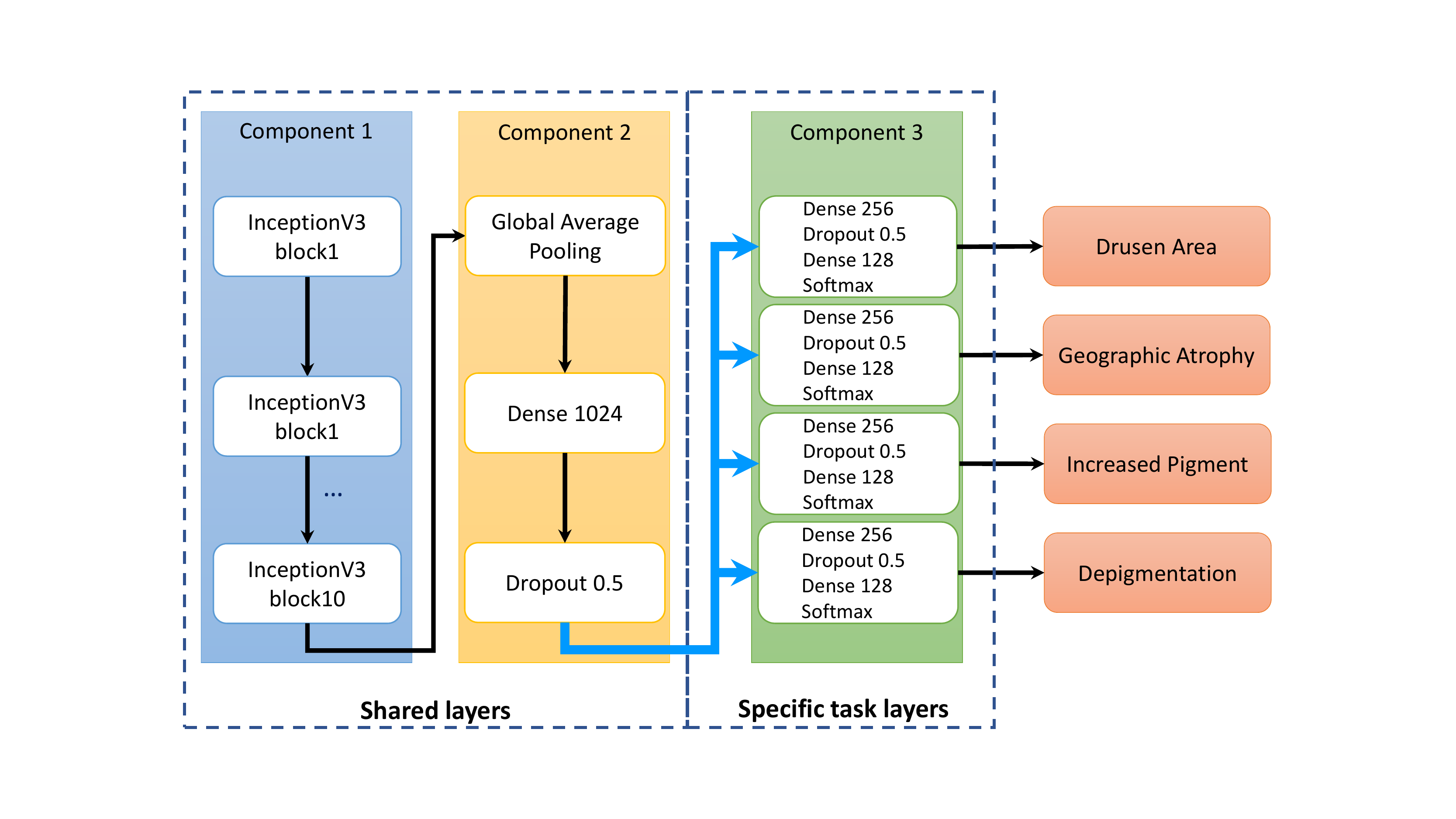}
\caption{The multi-task deep learning model for classification of age-related macular degeneration using the Age-Related Eye Disease Study Severity Scale.}
\label{fig:framework}
\vspace*{1em}
\end{figure}

The proposed method first grades the four characteristics (\ie, drusen area, geographic atrophy, increased pigment, and depigmentation), then calculates the overall AREDS Severity Scale score, based on the definitions described in Table~\ref{tab:amd}. The deep learning model for each AMD characteristic contains three components (Figure~\ref{fig:framework}). Component 1 consists of 10 Inception-V3 blocks, to capture image features\cite{szegedy2017inception}. On top of it, Component 2 consists of three layers: a global average pooling layer, a dense layer of dimension 1024, and a dropout layer. The global average pooling layer was applied to capture more informative features by enforcing the correspondence between features and classes\cite{lin2013network}. The dropout layer was used to reduce overfitting\cite{srivastava2014dropout}. Component 3 consists of four layers: a dense layer of dimension 256, a dropout layer, another dense layer of dimension 128, and a softmax layer; Component 3 is similar to the structure of our previous work. Components 1 and 2 form the basis of the transfer learning; in medical image classification, most methods use well-established architectures and add customized layers on top\cite{litjens2017survey}. Component 3 is used for multi-task learning, such that each individual task shares the common layers (Components 1 and 2) and has its own task-specific layers (Component 3).

\subsection*{Image pre-processing}

\begin{figure}[H]
\centering
\includegraphics[width=.8\textwidth,trim=0 13cm 17cm 0,clip]{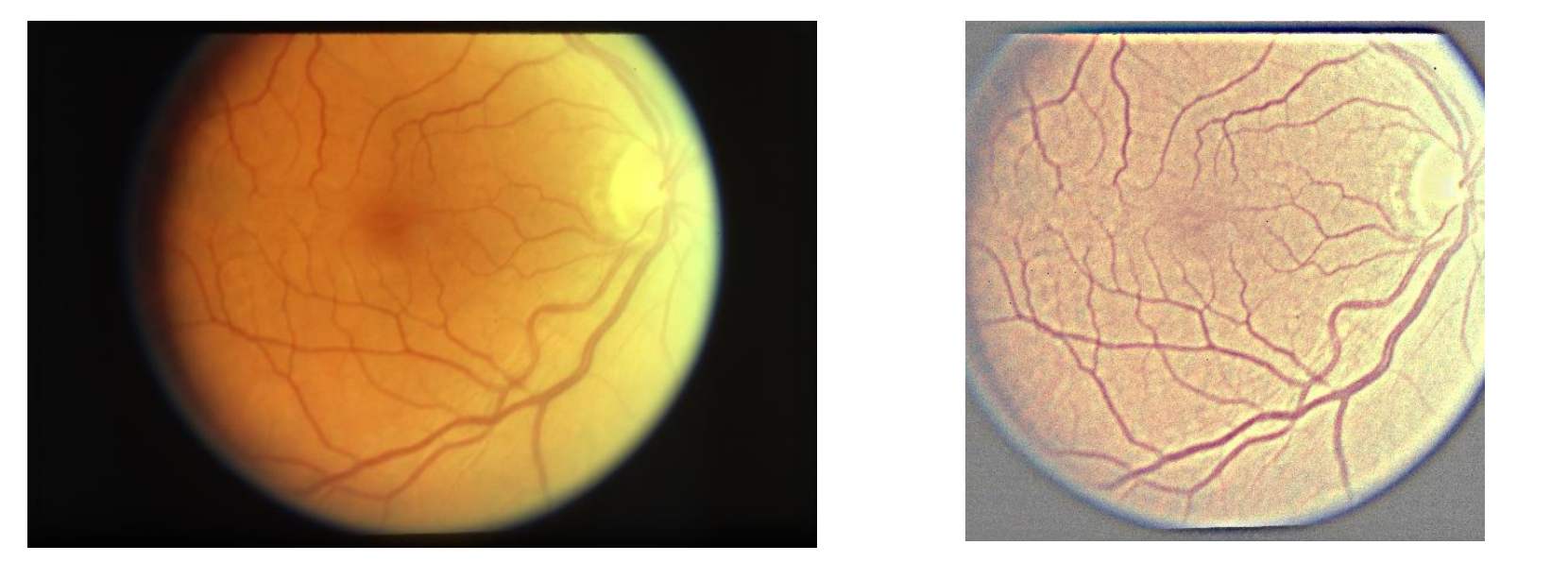}
\caption{An example of (a) the original image, and (b) the pre-processed color fundus photograph.}
\label{fig:preprocess}
\end{figure}

We manually examined the color fundus photographs and found that the images tended to have a relatively large area of black background, as well as large variations in brightness, which might adversely affect classification. For these reasons, image pre-processing was performed using three steps: (1) applying a Gaussian filter, with (0, 0) as Gaussian kernel size and (1000/30, 0) as Gaussian kernel standard deviation, to normalize the color balance (\ie{}, using similar methods to Grassmann \etal{}\cite{grassmann2018deep}), (2) cropping the images to a square shape, and (3) scaling the square images to 512$\times$512 pixels. Figure~\ref{fig:preprocess} shows an example of an image before and after pre-processing.

\subsection*{Training process}

We used data augmentation on the training set to enhance the robustness and generalizability of our model\cite{krizhevsky2012imagenet}. Specifically, each image in the training set was rotated randomly up to 180 degrees and flipped vertically or horizontally. 

We trained our model using the Adam optimizer\cite{kingma2015adam} with binary cross-entropy or categorical cross-entropy loss functions, and a mini-batch size of 16. For the Adam optimizer, we set the learning rate to 0.0001, beta1 to 0.9 and beta2 to 0.999. In addition, we applied dropout of 0.5 in the shared and task-specific layers. For weight initialization, pre-trained weights from the Inception-V3 network trained on the ImageNet dataset were used\cite{deng2009imagenet}. 

Since multi-task learning aims to reduce the losses of various tasks simultaneously, it requires a different training strategy compared to common single-task training. In this study, we used a two-phase approach for multi-task training\cite{pasunuru2017multi,liu2017adversarial,rao2018multi}. In the first phase, all the tasks were trained in parallel to obtain robust shared layers. In the second phase, each task was fine-tuned. In other words, the shared layers (Components 1 and 2 in Figure~\ref{fig:framework}) were blocked (\ie, set as non-trainable) and the task-specific layers (Component 3) were fine-tuned to achieve minimal loss. For each task, the model was saved based on the lowest loss on the validation set. Early stops were also applied to reduce overfitting. The training was stopped when the validation loss did not decrease in 10 steps during the first phase, and in 5 steps during the second phase.

All experiments were conducted on a server with 32 Intel Xeon CPUs, a NVIDIA GeForce GTX 1080 GPU, and a 512Gb of RAM. 

\subsection*{Datasets and Measurements}

The classification performance was evaluated using two different datasets of color fundus photographs (Table~\ref{tab:areds}), namely AREDS and AREDS2. The AREDS and AREDS2 were two large clinical trials sponsored by the National Eye Institute, National Institutes of Health\cite{areds1999age,chew2012age}. These two studies were designed to investigate the natural history and risk factors of AMD, as well as to evaluate the effects of nutritional supplementation on disease progression. The studies led to the development of the AREDS 9-step severity scale, based on color fundus photographs\cite{group2005age}. The scale was created and validated to classify AMD severity, primarily for research purposes, according to risk of progression to advanced AMD. Both datasets are held by a reading center located at the University of Wisconsin, Madison. The first image dataset (AREDS) contains approximately 60,000 color fundus photographs of eyes with non-advanced AMD and is publicly available on request\cite{areds1999age}. We used the AREDS dataset to train the model (\ie, the same dataset used by the state-of-art model\cite{grassmann2018deep}). The training, validation, and test sets comprised 64\%, 16\%, and 20\% of the AREDS patients, respectively. We used 5-fold cross-validation to measure the average performance of the models. The training, validation, and test sets were split at the level of individual patients (rather than individual eyes), such that all images from the same individual patient were kept in the same set. 

For the second image dataset (AREDS2), we randomly selected 1,500 color fundus photographs from the large database of AREDS2 images (which also contains other images, e.g. fundus autofluorescence images)\cite{chew2012age}. The sample distribution of the randomly selected images was similar to the population distribution in the whole dataset, and each randomly selected image was manually confirmed as a color fundus photograph (by a qualified ophthalmologist (TK)). We used the AREDS2 dataset as an independent test set for the evaluation of robustness and generalizability. 

Table~\ref{tab:areds} shows the distributions of the true AREDS 9-step scale from human grading at the Reading Center, separately for the two datasets.The distributions of AMD severity are quite different between the AREDS and AREDS2, since the inclusion criteria were distinct for the two clinical trials: the AREDS included participants with a wide spectrum of disease severity (including no AMD and early AMD), whereas the AREDS2 recruited participants with more severe disease only. Hence, the differences between the AREDS and AREDS2 datasets may be helpful in testing the generalizability of the model beyond the dataset on which it was trained.

For the measurements, standard machine learning classification metrics were reported: weighted precision, recall, F1-score, overall Kappa statistics, and accuracy.
\begin{table}[H]
\vspace{1em}
\centering
\small
\caption{Distributions of age-related macular degeneration severity, according to the AREDS severity scale (steps 1-9) in the AREDS and the AREDS2 image datasets.\label{tab:areds}}
\begin{tabular}{l@{\hspace{.4in}}rrr@{\hspace{.4in}}rr}
    \toprule
    AMD  & \multicolumn{2}{c}{AREDS} && \multicolumn{2}{c}{AREDS2}\\
    \cmidrule{2-3} \cmidrule{5-6}
    Severity Scale & n & \% && n & \%\\
    \midrule
    1 & 24,360 & 41.8 &&  11 &  0.7\\
    2 &  7,737 & 13.3 &&  11 &  0.7\\
    3 &  3,324 &  5.7 &&  23 &  1.5\\
    4 &  5,978 & 10.3 &&  66 &  4.2\\
    5 &  3,766 &  6.5 && 103 &  6.6\\
    6 &  4,641 &  8.0 && 353 & 22.6\\
    7 &  3,953 &  6.8 && 650 & 41.6\\
    8 &  3,396 &  5.8 && 225 & 14.4\\
    9 &  1,130 &  1.8 && 120 &  7.7\\
    \textit{Total} & 58,285 &  && 1,562 & \\
    \bottomrule
\end{tabular}
\end{table}

\section*{Results}

In addition to the multi-task learning model described above (`\multitaskmodel{}', as shown in Figure~\ref{fig:framework}), two additional models were evaluated, for comparison: the Convolutional Neural Network (CNN) model developed by Grassmann \etal\cite{grassmann2018deep}(`\grassmannmodel{}'), and our CNN model (`\cnnmodel{}'). The reason we implemented our own \cnnmodel{} is because the authors of the original \grassmannmodel{} made available only the weights (trained using the AREDS dataset), but not the source code and data partition. Since our test set may have overlapped with the training set used in the study of Grassmann \etal\cite{grassmann2018deep}, it is difficult to perform a direct comparison between our \cnnmodel{} and \grassmannmodel{} on the AREDS dataset. 

The \multitaskmodel{} and the \cnnmodel{} were tested on the AREDS dataset. Their performances are shown in Table~\ref{tab:5-fold}. In all metrics except precision, the performance of the Multi-task model was superior to that of the \cnnmodel{}. All three models (the \multitaskmodel{}, the \cnnmodel{}, and the \grassmannmodel{}) were tested using the AREDS2 dataset. Their performances are also shown in Table~\ref{tab:5-fold}. Again, the performance of the \multitaskmodel{} was superior to that of the \cnnmodel{}, with a higher degree of superiority observed in the AREDS2 dataset than in the AREDS dataset. In particular, the performance was approximately 5\%, 3\%, and 4\% higher, in terms of weighted F1-score, overall Kappa statistics, and overall accuracy, respectively. Since the AREDS2 dataset was an independent dataset not used for training the models, these results suggest that the \multitaskmodel{} may be more robust and generalizable than the traditional CNN model. In addition, the \grassmannmodel{} had more than 10\% lower accuracy than either the \multitaskmodel{} or the \cnnmodel{} (discussed further below).
\begin{table}[H]
\vspace*{1em}
\small
\centering
\caption{5-fold cross validation performance of the proposed \multitaskmodel{}, the \cnnmodel{}, and the \grassmannmodel{}, on the AREDS and AREDS2 datasets.\label{tab:5-fold}}
\vspace*{-1em}
\begin{tabular}{l@{\hspace{.4in}}ccc@{\hspace{.3in}}ccc}
    \toprule
              & \multicolumn{2}{c}{AREDS} && \multicolumn{3}{c}{AREDS2}\\
    \cmidrule{2-3}\cmidrule{5-7}
              & Multi-task & CNN       && Multi-task & CNN & Grassmann\\
    \midrule
    Precision & 0.591 & \myemph{0.592} && \myemph{0.545} & 0.533 & 0.514\\
    Recall    & \myemph{0.614} & 0.612 && \myemph{0.472} & 0.437 & 0.345\\
    F1-score  & \myemph{0.597} & 0.593 && \myemph{0.492} & 0.449 & 0.398\\
    Kappa     & \myemph{0.487} & 0.484 && \myemph{0.330} & 0.297 & 0.216\\
    Accuracy  & \myemph{0.614} & 0.612 && \myemph{0.472} & 0.437 & 0.346\\
    \bottomrule
\end{tabular}
\end{table}

Table~\ref{tab:1-fold} compares the classification results for multi-task learning and single-task learning on models trained from one of the 5-fold cross-validations. Single-task learning means that there are effectively four CNN models, each of which is trained independently. The results show that multi-task learning achieved better performance on all the metrics (weighted F1-score, overall Kappa statistics and overall accuracy) in both datasets. In particular, it had $\sim4\%$ higher weighted F1-score and $\sim2\%$ higher for other metrics.
\begin{table}[H]
\vspace*{1em}
\small
\centering
\caption{1-fold performance of multi-task and single-task learning on the AREDS and AREDS2 datasets.\label{tab:1-fold}}
\vspace*{-1em}
\begin{tabular}{l@{\hspace{.3in}}ccc@{\hspace{.3in}}cc}
    \toprule
             & \multicolumn{2}{c}{AREDS} && \multicolumn{2}{c}{AREDS2}\\
    \cmidrule{2-3}\cmidrule{5-6}
             & Multi-task & Single-task && Multi-task & Single-task\\
    \midrule
    F1-score & \myemph{0.621} & 0.583 && \myemph{0.502} & 0.483\\
    Kappa    & \myemph{0.485} & 0.466 && \myemph{0.340} & 0.312\\
    Accuracy & \myemph{0.621} & 0.607 && \myemph{0.480} & 0.461\\
    \bottomrule
\end{tabular}
\vspace*{-1em}
\end{table}

\section*{Discussion}

The above results demonstrate that our \multitaskmodel{} achieved superior performance to the state-of-the-art model in the two evaluation datasets. In this section, we consider potential areas for improvement; in particular, we performed error analysis on four individual risk factors and comparatively analyzed the difference between evaluation datasets. Additionally, we also quantitatively illustrate that transfer learning improves the generalization capability of models.

\subsection*{Comparisons between four characteristics}

\begin{figure}[H]
\begin{minipage}[t]{0.36\textwidth}
\includegraphics[width=\textwidth]{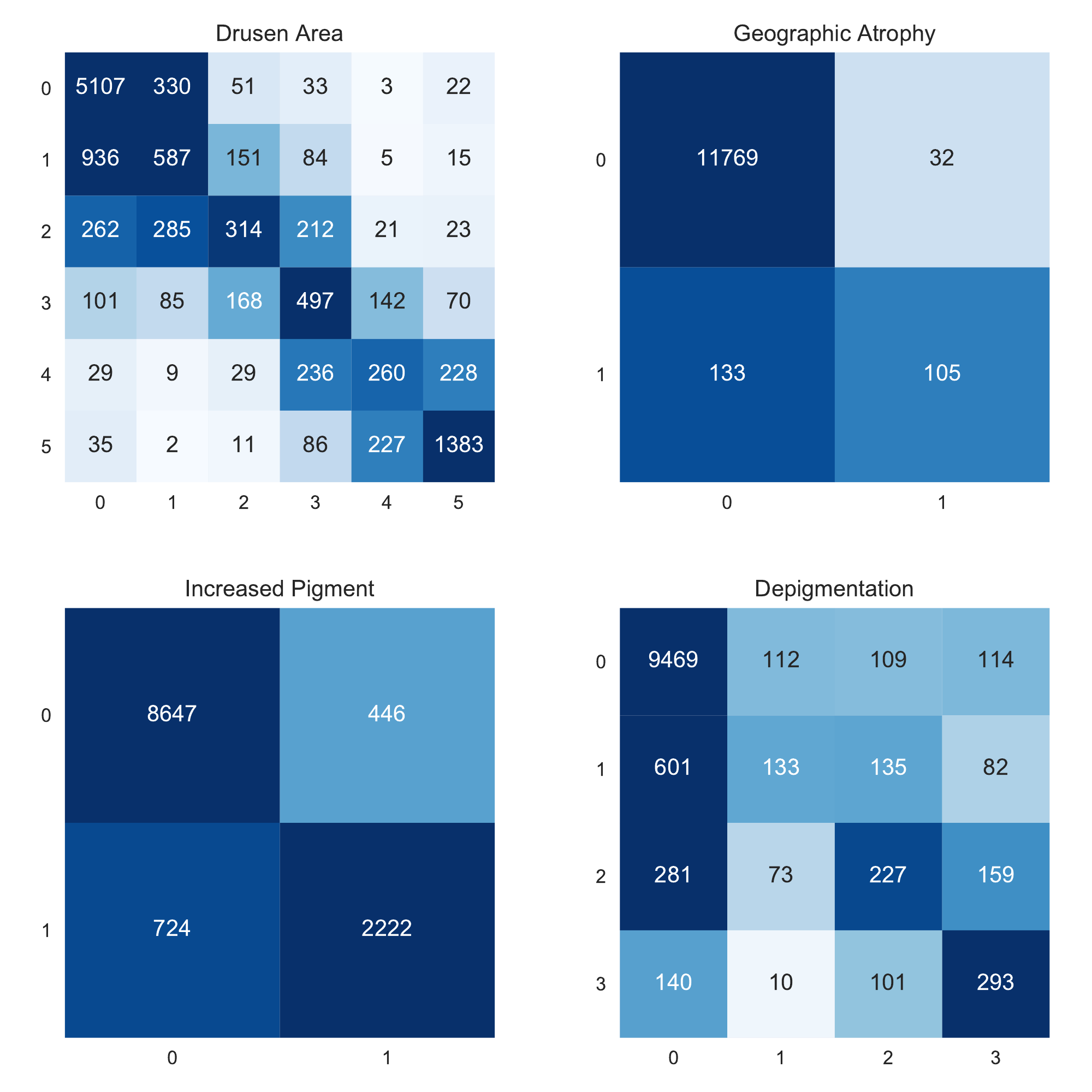}
\vspace*{-2em}
\caption{Confusion matrices for individual AMD characteristics.}
\label{fig:confusion}
\end{minipage}
\hspace*{1em}
%
\begin{minipage}[t]{0.6\textwidth}
\centering
\includegraphics[width=\textwidth,clip,trim=2cm 2cm 2cm 2cm]{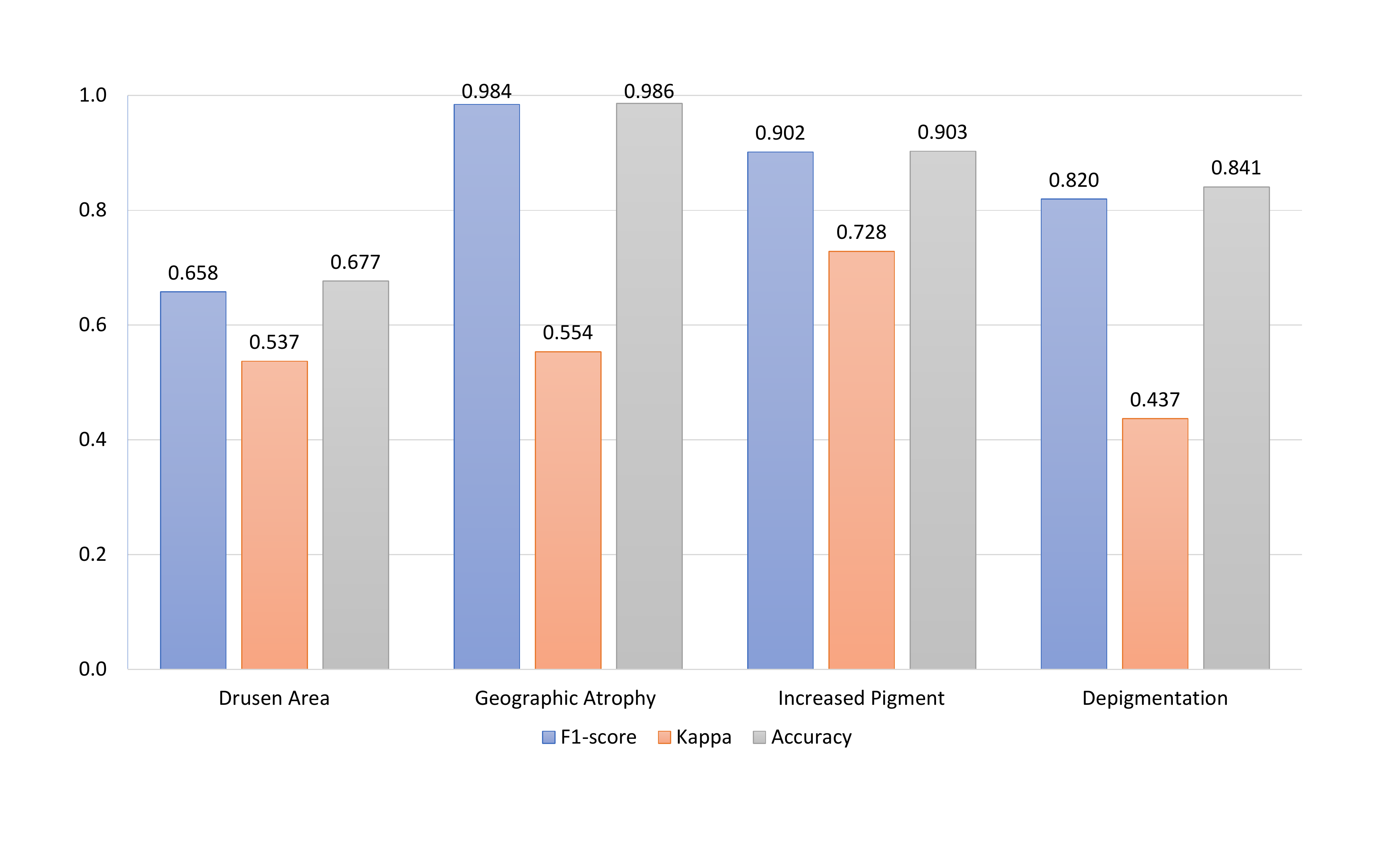}
\vspace*{-2em}
\caption{Performance of individual AMD characteristics.}
\label{fig:individual}
\end{minipage}
\end{figure}

Since four characteristics contribute to the final AREDS severity scale, we analyzed the performance of the \multitaskmodel{} separately for each characteristic. Figure~\ref{fig:confusion} shows the confusion matrices and Figure~\ref{fig:individual} shows the overall performance of these characteristics. 

We made three observations. First, as regards the performance of the model for each of the four characteristics, its performance in correctly grading drusen area had substantially lower F1-score and accuracy than those for the other three characteristics (\eg{}, accuracy of 0.68 for drusen area versus 0.99, 0.90, and 0.84 for geographic atrophy, increased pigment, and depigmentation, respectively). Indeed, its accuracy in grading drusen area was similar to its overall accuracy in predicting the AREDS severity scale (0.68 and 0.62, respectively). This suggests that grading of drusen area may be the limiting factor in the overall performance of the model, such that future improvements in this particular task may contribute most to increasing the model's performance.

Second, at the class level, some classes have substantially higher levels of misclassification than others (as observed in the confusion matrices). For example, the accuracies of drusen area classification for classes 1, 2, 3, and 4 were 0.33, 0.28, 0.47, and 0.33, respectively. These are much lower than the accuracies for classes 0 and 5, which were 0.92 and 0.79, respectively. This is likely because the differences between classes 2, 3, and 4 (small drusen) were not large enough for effective classification, whereas class 0 (no drusen) and class 5 (large drusen) represent more distinct categories that may be easier for classification. Similarly, classes 1 and 2 for depigmentation also have low performance: the macro accuracies were 0.14 and 0.31, respectively.

Third, at the data level, the number of instances is still limited. As observed in the confusion matrices, for all four individual characteristics, all classes except 0 have a limited number of instances. It is likely that these relatively small numbers of instances may be insufficient for models to learn. While both data augmentation (which implicitly increased the number of instances) and transfer learning were applied in our model to address this problem, it would still be valuable to increase the number of instances further (\ie, new images from additional patients), in order to improve performance by distinguishing more accurately between classes. Indeed, data imbalance is a common problem for medical image classification\cite{litjens2017survey}. However, traditional sub-sampling and over-sampling strategies may not be applicable in this case. Sub-sampling would limit the number of instances of majority classes, while over-sampling would require more distinct instances.

Taken together, we consider that (i) instead of reusing models developed for general computer vision purposes, domain-specific models are needed, and (ii) it is pressing to have annotation of more instances.


\subsection*{Comparisons between the AREDS and AREDS2}

While the \multitaskmodel{} was more robust and generalizable than the other models tested, the performance of all the models decreased substantially from testing on the AREDS dataset to the AREDS2 dataset. The main reason for this is likely to be the substantial differences between their class distributions. For instance, as shown in Table~\ref{tab:areds}, the AREDS dataset contains a large proportion of images with score 1, whereas the AREDS2 dataset contains an extremely small proportion of these. Indeed, skewed distributions make many deep learning algorithms less effective, especially in predicting minority class examples. 

\begin{figure}
\centering
\includegraphics[width=.65\textwidth,clip,trim=2cm 3cm 2cm 2cm]{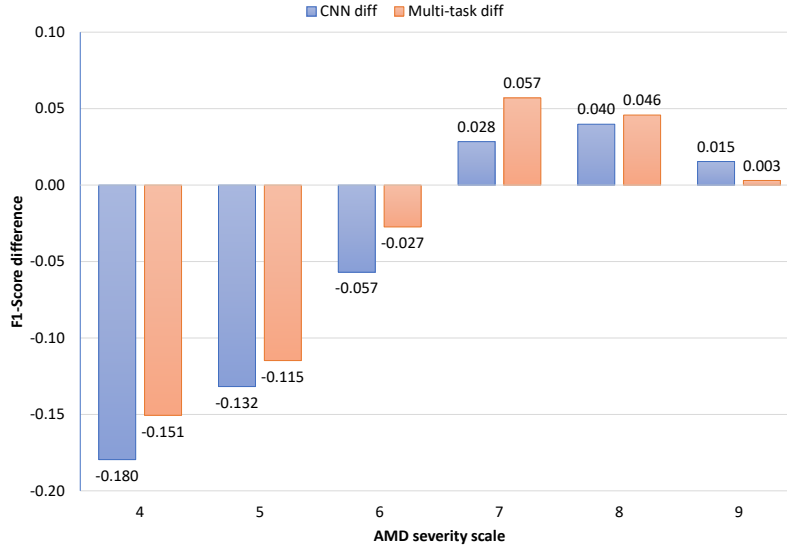}
\vspace*{-1em}
\caption{F1-score class difference (AREDS vs AREDS2). For example, the CNN model had 17.96\% lower F1-score on class 4 compared AREDS to AREDS2.}
\label{fig:diff}
\end{figure}

For the \multitaskmodel{} and the \cnnmodel{}, we analyzed the differences in F1-score, Kappa, and accuracy (separately for each class) between the AREDS and AREDS2 datasets (Figure~\ref{fig:diff}). Here, we selected steps 4--9 for statistical analysis, because these scores had more than 50 instances in each dataset. In terms of altered performance between testing on the AREDS and AREDS2 datasets, the \multitaskmodel{} was superior to the \cnnmodel{} for 5 out of the 6 steps. For steps 4--6, the \multitaskmodel{} suffered from less loss in performance than the \cnnmodel{}; for example, for step 4, the F1-score of the \cnnmodel{} decreased by 18\% , while that of the \multitaskmodel{} decreased by 15\%. In addition, for steps 7-8, the \multitaskmodel{} demonstrated higher performance gains between testing on the AREDS and AREDS2 dataset; for example, for scale 7, the F1-score of the \cnnmodel{} increased by 3\% , whereas that of the \multitaskmodel{} increased by 6\%. This analysis shows that the \multitaskmodel{} not only had superior performance overall, but also achieved superior performance for most of the classes.

Overall, the substantial decrease in performance between the AREDS and AREDS2 datasets suggests several points: (i) in principle, for medical image classification, models should favor high robustness and generalizability, (ii) for transfer learning, fine-tuning may achieve superior results than feature extraction, because of distribution differences in various medical image datasets, and (iii) it is of value to accumulate different medical datasets. In this study, the AREDS and AREDS2 datasets complement one another, such that combining them would provide a larger and more balanced overall dataset.

The accuracy of the \cnnmodel{} released by Grassmann \etal was more than 10\% lower than our CNN implementation~\cite{grassmann2018deep}. One important factor might be that the \grassmannmodel{} did not use pre-trained model weights, \ie, weights trained from millions of images taken from general domains. Instead, the \grassmannmodel{} was trained from scratch using a random weight initializer. While the AREDS dataset contains a relatively large number of images, the distribution is unbalanced. Thus, the model may not have learnt the minority classes effectively and may suffer from overfitting. To investigate this hypothesis, we compared the performance of the \emph{CNN+random model} (the model without using pre-trained weights) and the \emph{CNN+pre-trained model} (the model using pre-trained weights). Table~\ref{tab:pretrained} shows that the performance of the two models is relatively similar on testing using the AREDS dataset; however, on testing using the AREDS2 dataset, the performance of the \emph{CNN+random model} was substantially lower than that of the \emph{CNN+pretrained model}, with a difference of 8\% in F1-scores. This finding is consistent with a previous study that examined the performance of full-training and fine-tuning using pre-trained weights in three medical image applications\cite{tajbakhsh2016convolutional}. This study demonstrated that using pre-trained weights was associated not only with superior performance but also with increased generalizability. 
\begin{table}[H]
\vspace*{1em}
\small
\centering
\caption{The performance between the CNN model using pretrained weights and randomly initialized weights on AREDS and AREDS2.\label{tab:pretrained}}
\vspace*{-1em}
\begin{tabular}{l@{\hspace{.3in}}ccc@{\hspace{.3in}}cc}
    \toprule
    & \multicolumn{2}{c}{AREDS} && \multicolumn{2}{c}{AREDS2}\\
    \cmidrule{2-3}\cmidrule{5-6}
             & Pretrained & Random && Pretrained & Random\\
    \midrule
    F1-score & \textbf{\textit{0.588}} & 0.582 && \textbf{\textit{0.487}} & 0.406\\
    Kappa    & \textbf{\textit{0.476}} & 0.470 && \textbf{\textit{0.338}} & 0.258\\
    Accuracy & \textbf{\textit{0.615}} & 0.610 && \textbf{\textit{0.472}} & 0.391\\
    \bottomrule
\end{tabular}
\vspace{-1em}
\end{table}

\section*{Conclusion}

In conclusion, we have developed a new deep learning model for AMD classification of the AREDS severity scale. Evaluation on two datasets showed that our model was consistently superior to the current state-of-the-art model. In addition, our model was able to classify four separate components of the AREDS Severity Scale, thus providing improved transparency. Error analysis revealed that the limiting factor of our model’s performance was in the classification of the drusen area component. In the future, we plan to improve our model by concentrating on drusen area classification by obtaining datasets with a higher proportion of non-advanced AMD cases.

\section*{Acknowledgements}
This work was supported by the Intramural Research Programs of the National Institutes of Health, National Library of Medicine and the National Eye Institute. We also thank Daniel Li for his help.

\makeatletter
\renewcommand{\@biblabel}[1]{\hfill #1.}
\makeatother

\bibliographystyle{vancouver}
\setlength\bibitemsep{4pt}
\bibliography{bib}
\end{document}